\documentclass[letterpaper, 10 pt, conference]{ieeeconf}  

\IEEEoverridecommandlockouts                              

\usepackage{amsmath} 

\usepackage{graphicx}
\usepackage{todonotes}

\usepackage[hidelinks]{hyperref}

\usepackage{booktabs} 
\usepackage{array}    
\newcolumntype{L}[1]{>{\raggedright\arraybackslash}p{#1}} 
\usepackage{subcaption} 

\title{\LARGE \bf
A Sensorless, Inherently Compliant Anthropomorphic Musculoskeletal Hand Driven by Electrohydraulic Actuators
}

\author{Misato Sonoda$^{1, 2}$,
Ronan Hinchet$^{1}$,
Amirhossein Kazemipour$^{1}$,
Yasunori Toshimitsu$^{1}$, 
Robert K. Katzschmann$^{1}$
\thanks{$^{1}$Soft Robotics Lab, ETH Zurich, Switzerland}
\thanks{$^{2}$The University of Tokyo, Japan}
\thanks{\tt\small \href{mailto:m-sonoda@jsk.imi.i.u-tokyo.ac.jp}{m-sonoda@jsk.imi.i.u-tokyo.ac.jp}, \{\href{mailto:rhinchet@ethz.ch}{rhinchet}, \href{mailto:akazemi@ethz.ch}{akazemi}, \href{mailto:ytoshimitsu@ethz.ch}{ytoshimitsu}, \href{mailto:rkk@ethz.ch}{rkk}\}@ethz.ch}
}

\begin{document}

\maketitle
\thispagestyle{empty}
\pagestyle{empty}

\begin{abstract}

Robotic manipulation in unstructured environments requires end-effectors that combine kinematic dexterity with physical compliance.
While traditional rigid hands rely on complex external sensors for safe interaction, electrohydraulic actuators offer a promising alternative by combining muscle-like compliance with self-sensing capability.
This paper presents the design, control, and evaluation of a musculoskeletal robotic hand architecture powered entirely by remote Peano-HASEL actuators, optimized for safe manipulation.
By relocating the actuators to the forearm, we isolate the grasping interface from electrical hazards while maintaining a slim, human-like profile.
To address the inherently limited linear contraction of these soft actuators, we integrate a 1:2 pulley routing mechanism that mechanically amplifies tendon displacement.
The resulting system prioritizes compliant interaction over high payload capacity, leveraging the intrinsic force-limiting characteristics of the actuators to provide inherent safety.
This physical safety is augmented by the self-sensing nature of the HASEL actuators: by monitoring the operating current alone, we achieve real-time grasp detection and closed-loop contact-aware control without external force transducers or encoders.
Experimental results demonstrate the system's dexterity and safety through the execution of grasp types from standard taxonomies and the non-destructive grasping of highly fragile objects such as a paper balloon.
These findings represent a step toward simplified, inherently compliant soft robotic manipulation.

\end{abstract}

\section{Introduction}\label{sec:introduction}

Robotic manipulation in unstructured environments demands end-effectors that possess both kinematic dexterity and physical compliance.
Traditional rigid robotic hands, typically driven by electromagnetic motors, excel in precision and payload capacity.
However, they lack inherent physical compliance, and therefore rely on complex and expensive external sensors \cite{shadow_hand, dla_hand2, sdm_hand}, such as multi-axis force/torque sensors or tactile arrays, to ensure safe interactions with fragile objects, unknown shapes, or humans.
This reliance on sensors increases the system's complexity, weight, and computational overhead.

\begin{figure}[!h]
\vspace{0.5em}
\centering
\includegraphics[width=\columnwidth]{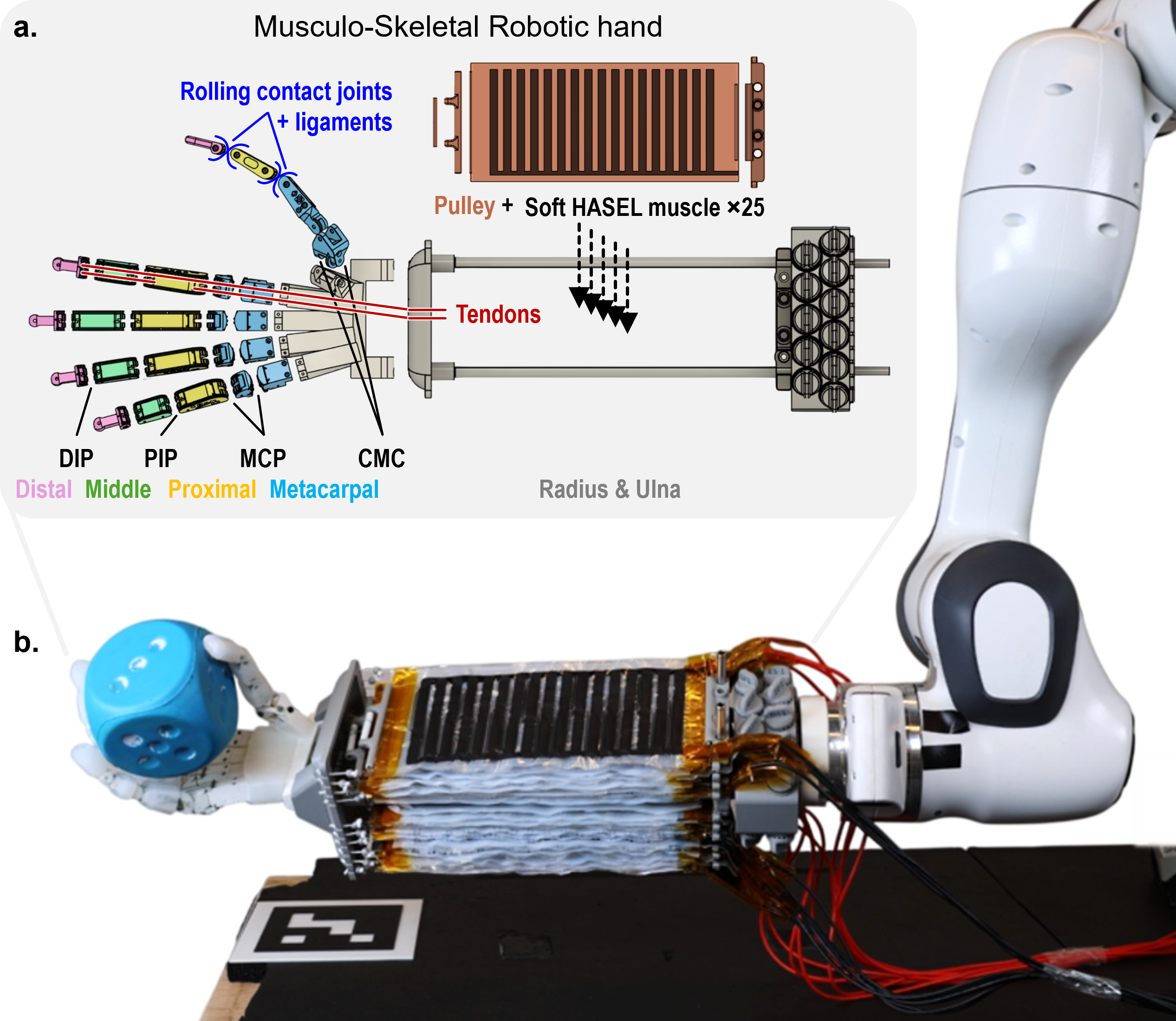}
\caption{\textbf{A tendon-driven robotic hand system actuated with HASEL actuators.} (a) Soft HASEL muscle actuators in the forearm drive the fingers via tendon transmission and pulley mechanisms, enabling compliant finger motion through rolling-contact joints. (b) The proposed hand system is grasping a cube. All fingers can be controlled independently.}
\vspace{-1em}
\label{fig:system_overview}
\end{figure}

Soft robotics has emerged as a compelling alternative, using compliant materials and structures to naturally adapt to object geometries and absorb external impacts.
While fluid-driven soft actuators offer excellent compliance \cite{rbohand2, pneunets_gripper}, they typically require bulky compressors, valves, and tubes.
Recently, electrohydraulic actuators, specifically Hydraulically Amplified Self-healing ELectrostatic (HASEL) actuators, have gained attention \cite{hasel_science, hasel_actuators}.
They combine the fast response times of dielectric elastomer actuators with the versatility of fluidic systems, offering muscle-like actuation that is inherently compliant and capable of self-sensing, as reported in several robotic applications \cite{miniaturized_circuitry, hasel_finger, pele_leg, hasel_shoulder}.

Despite these advantages, integrating HASEL actuators into highly dexterous robotic hands remains a challenge.
Most existing HASEL-based grippers \cite{hasel_science, hasel_gripper} rely on the direct bending or deformation of the actuators themselves.
While suitable for simple pinching or wrapping motions, this direct-actuation approach limits the grasp taxonomy and kinematic precision compared to anthropomorphic skeletal designs driven by tendons.
Furthermore, the contraction strain of HASEL actuators is typically limited to below 15\,\% \cite{hasel_analytical_model}, which is often insufficient to fully drive the large joint excursions of human-inspired hand designs.
Placing high-voltage actuators directly at the grasping interface also raises safety concerns and increases the bulk of the fingers.

Recent work has shown the potential of HASEL-driven prosthetics to overcome some of these kinematic limits.
Most notably, the high-speed prosthetic finger presented in \cite{hasel_finger} achieved fingertip forces exceeding $1$\,N and a range of motion of $77^\circ$ by coupling seven HASEL actuators in parallel to pull the base of a rigid four-bar linkage.
However, this rigid linkage and parallel actuation strategy was dedicated to a single digit: directly replicating it for all fingers of an anthropomorphic hand would result in bulk, complex routing, and a loss of the inherent physical compliance desired in soft robotics.
Scaling to a fully integrated, multi-fingered hand therefore requires optimizing system-level integration, space efficiency, and interactive safety rather than the raw power of an isolated digit.

To address these challenges and extend the HASEL-driven prosthetics approach, this paper presents an integrated musculoskeletal robotic hand architecture powered entirely by remote Peano-HASEL actuators.
Our approach bridges the gap between soft-actuator compliance and rigid kinematic dexterity by combining a remote actuation paradigm with a compliant, human-inspired transmission.
Building on a tendon-driven architecture, we relocate the HASEL actuators to a human-sized forearm, isolating the grasping interface from electrical hazards while maintaining a slim, human-like hand profile.
To overcome the inherent stroke limitations of the soft actuators without resorting to massive parallel stacks, we implement a 1:2 pulley routing mechanism that mechanically amplifies tendon displacement, targeting the full range of motion of the rolling-contact finger joints.
Crucially, the design is optimized for the safe manipulation of fragile objects: rather than pursuing high payload capacity, the architecture prioritizes compliant interaction by leveraging the intrinsic force-limiting characteristics of HASEL actuators.
This ensures non-destructive contact with delicate items and environments, providing inherent safety without the need for complex, high-bandwidth active force control.

The main contributions of this paper are threefold:
\begin{itemize}
    \item \textbf{Hand architecture.} A holistic design of an anthropomorphic robotic hand optimized for safe manipulation. By combining a remote-actuation paradigm with a stroke-amplifying transmission, the architecture achieves a slim, human-like form factor while targeting the kinematic range needed for dexterous manipulation.
    \item \textbf{Sensorless control.} A manipulation approach that leverages the capacitive nature of HASEL actuators through operating-current monitoring. This enables real-time grasp detection and safe, closed-loop control for delicate grasping without external encoders or force sensors.
    \item \textbf{Experimental demonstration.} A demonstration of the system's dexterity and inherent safety through the execution of grasp postures from standard taxonomies and the non-destructive manipulation of extremely fragile objects such as a paper balloon.
\end{itemize}

\section{Robotic Hand Mechanism}

\subsection{Actuator Design and Characterization} \label{subsec:actuator_design}

Peano-HASEL actuators \cite{hasel_actuators} were fabricated as follows: the basic structure consists of a flexible pouch made from 25\,\textmu m Mylar (PET) films. Carbon-ink electrodes were screen-printed onto the films and cured. The pouches were then filled with a liquid dielectric (silicone oil, 5\,cSt) and heat-sealed.

\begin{figure}[tb]
\centering
\includegraphics[width=\columnwidth]{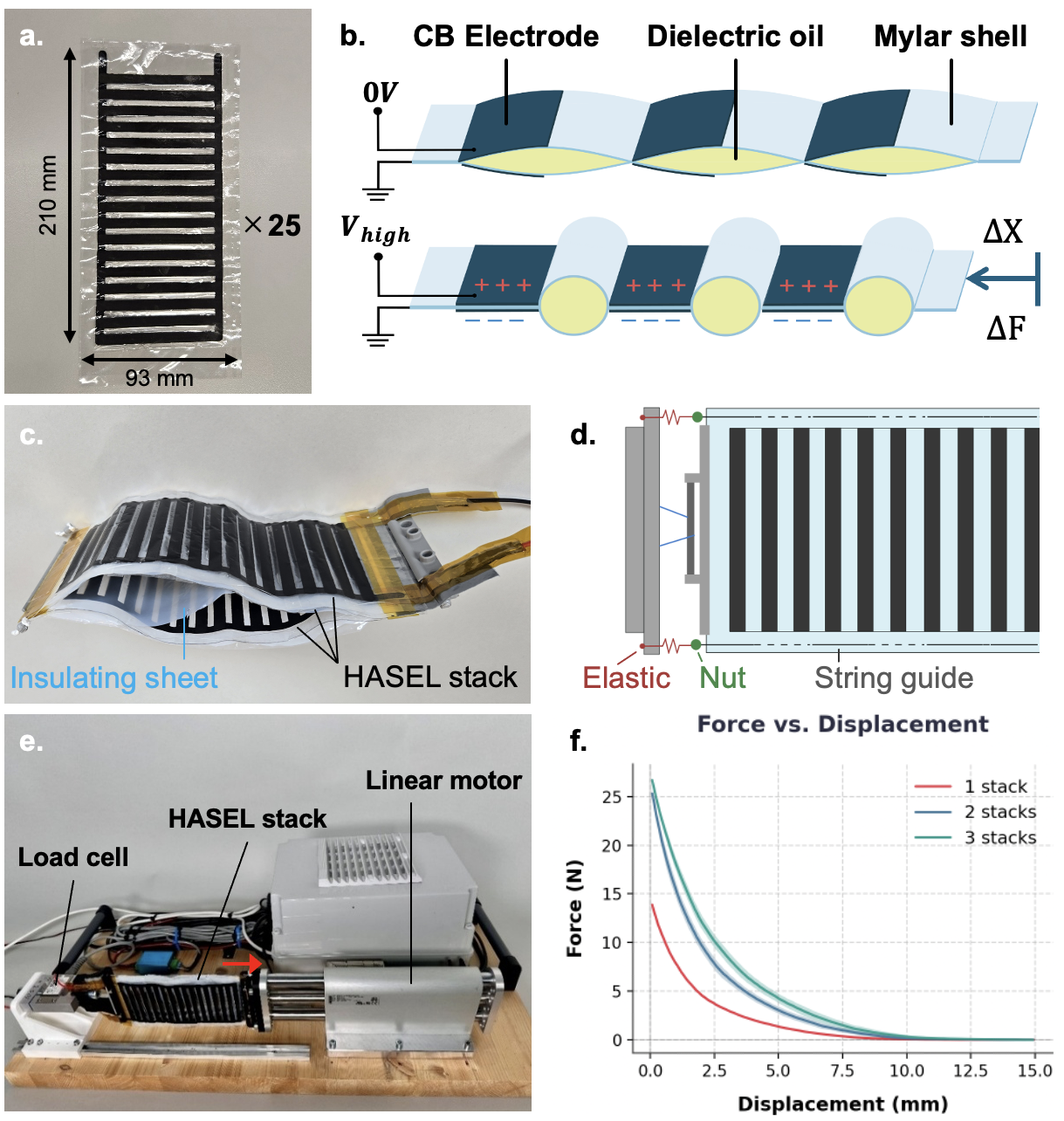}
\caption{\textbf{Soft Muscle Stacking.} (a) A fabricated Peano-HASEL actuator measures 210\,mm by 93\,mm. (b) Actuation principle of the Peano-HASEL actuator proposed in \cite{hasel_actuators}. (c) Connected actuator stack forming a soft muscle. (d) Guiding structure of the actuators. (e) Force-displacement characterization setup using a linear motor and a load cell. (f) Force-displacement characteristics of single, double, and triple-stacked Peano-HASEL configurations with the shaded standard deviation.}
\label{fig:actuator_overview}
\end{figure}

To amplify the output force for the robotic hand, multiple Peano-HASEL actuators (Fig.~\ref{fig:actuator_overview}(a)) were connected in parallel to form actuator stacks.
To ensure safe operation at high voltage and prevent electrical arcing or short circuits, silicone spray was applied to the outer surfaces of the actuators.
In addition, 50\,\textmu m PTFE sheets were interleaved between each actuator in the stack; these serve the dual purpose of providing electrical insulation and minimizing inter-actuator friction during contraction.

To determine the number of parallel actuators required for each finger joint, the force-displacement characteristics of the stacks were evaluated using a custom pull tester with a linear motor (C1250-MI-XC-0S-000, LinMot) and a load cell (Model 614, Tedea Huntleigh).
Under a $5.5$\,kV operating voltage, the linear motor stretched the actuator stacks by $15$\,mm at a constant velocity of $0.5$\,mm/s, continuously recording the contraction force over $5$ strokes for each configuration.
The resulting force-displacement curves for single ($10$\,g), double ($22$\,g), and triple-stack ($34$\,g) configurations are compared in Fig.~\ref{fig:actuator_overview}(f).
Adding parallel units enhances the output force but also adds weight and bulk to the forearm.
The stack configuration for each joint was therefore chosen by balancing the kinematic and task requirements of the finger against the overall weight and space constraints of the system.


In the hand design, the distal and proximal interphalangeal (DIP/PIP) joints require a maximum tendon excursion of approximately $12$\,mm to reach $90^\circ$ flexion.
To accommodate the limited stroke of the HASEL actuators, which deliver usable force only up to ${\sim}10$\,mm (Fig.~\ref{fig:actuator_overview}(f)), we integrated a 1:2 pulley routing system on top of the actuator stacks.
This ratio is constraint-derived: while the joints require $12$--$17$\,mm excursions, the 1:2 amplification maps these requirements onto manageable $6$--$8.5$\,mm actuator strokes. 
While higher ratios would excessively attenuate tendon force, this pulley configuration preserves transmission compliance and adds little structural volume, unlike rigid linkages \cite{hasel_finger}.
For these joints, the two-stack configuration provided sufficient contraction force across the $6$\,mm operating range ($25.3$\,N at the initial position decreasing to $2.0$\,N at $6$\,mm contraction) while keeping the system lightweight.

In contrast, the metacarpophalangeal (MCP) joints require a larger tendon excursion of up to $17$\,mm, which translates to an $8.5$\,mm actuator contraction through the pulley system, and typically bear higher mechanical loads during power grasping. 
The heavier three-stack configuration is therefore selected to maximize the available force over this longer stroke within the weight and space budget of the forearm.
This setup provides a contraction force ranging from $26.1$\,N initially to $0.8$\,N at full $8.5$\,mm excursion.
Note that the contraction force diminishes considerably toward the end of the stroke (Fig.~\ref{fig:actuator_overview}(f)), leaving a limited force reserve near full excursion; this is consistent with the reduced range of motion observed in Fig.~\ref{fig:characterization}.

\subsection{Robotic Hand Design} \label{subsec:hand_design}

The mechanical design of the hand is based on a circular rolling-contact joint \cite{collins2003kinematics,faive_hand} with a dynamically shifting center of rotation to ensure space-efficient flexion.

The index, middle, ring, and pinky fingers consist of three rigid phalanges (proximal, intermediate, and distal) connected by rolling-contact joints with ligaments at the metacarpophalangeal (MCP), proximal interphalangeal (PIP), and distal interphalangeal (DIP) joints, allowing $90^\circ$ joint flexion and $30^\circ$ abduction.
While the MCP joint is driven independently by flexor and extensor tendons, the PIP and DIP joints are mechanically coupled to flex and extend synergistically, as in a human hand.

To achieve human-like opposition, the thumb incorporates rolling-contact interphalangeal (IP) and MCP joints combined with a pin-joint carpometacarpal (CMC) joint; the CMC abduction is fixed in this prototype, providing two active degrees of freedom (DOF). 
The rigid palm provides structural support for the fingers and routes the tendons toward the forearm.
To enhance grasping performance and safety, the palm and finger surfaces are enclosed in a custom-molded soft silicone skin (Dragon Skin~10, Smooth-On), which improves grip and helps prevent object slippage.

\begin{figure}[tb]
\centering
\includegraphics[width=\columnwidth]{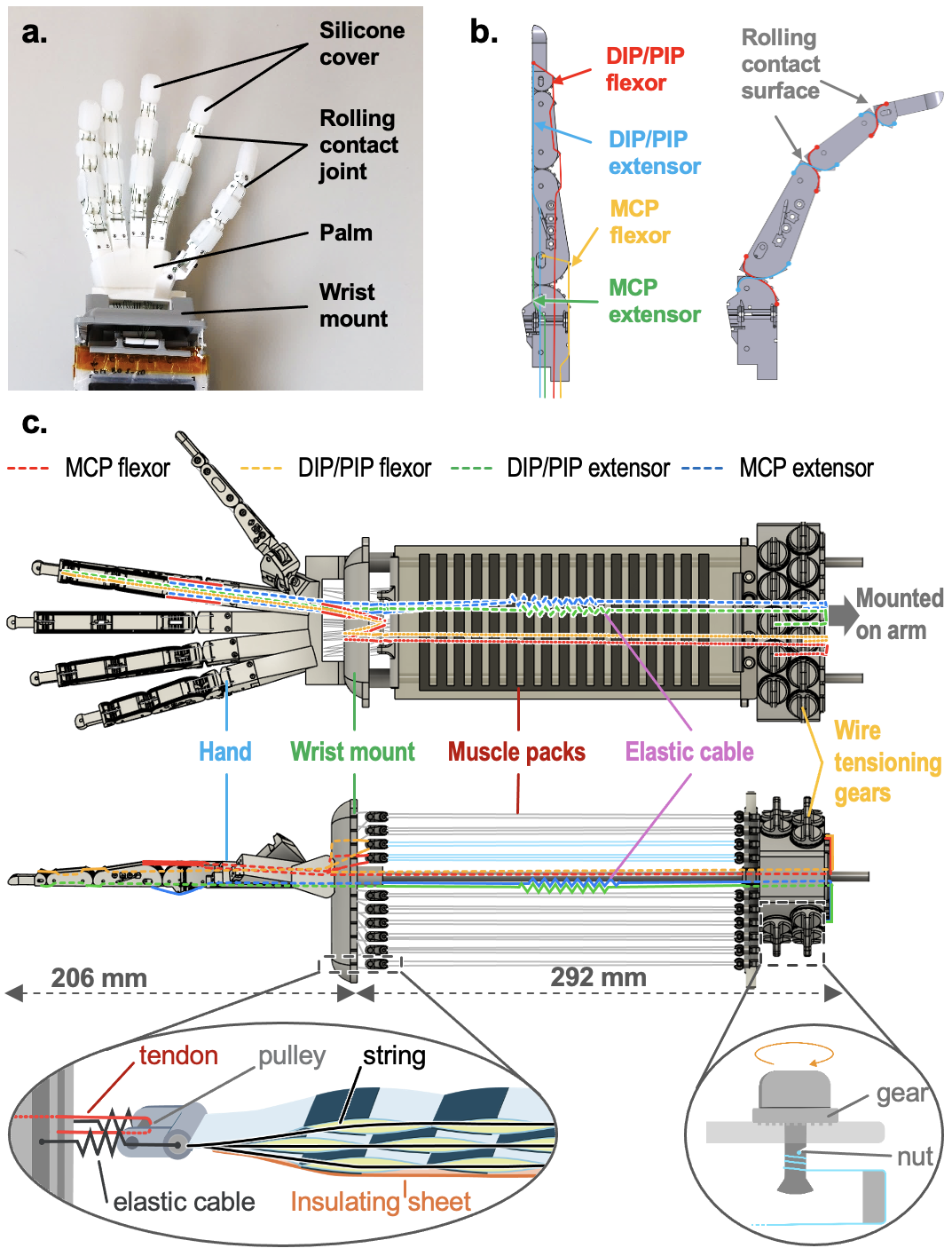}
\caption{\textbf{Tendon-Driven Hand Design.} (a) Overview of the modular robotic hand. (b) Flexor and extensor tendon routing of the index finger, demonstrating the rolling-contact joints and coupled PIP/DIP mechanism. (c) Actuator packs are stacked in parallel in the forearm (e.g., 2-stack for DIP/PIP, 3-stack for MCP). The flexor tendons utilize a pulley system attached to the actuators, while the extensor tendons are connected in series with elastic cords for passive restoration. All tendons are anchored and adjusted at the wire tensioning gears located at the bottom.}
\label{fig:hand_design}
\end{figure}

\subsection{Transmission Mechanism} \label{subsec:transmission_mechanism}

To convert the linear contraction of the HASEL actuators into independent finger movements, the hand uses a flexor--extensor tendon routing architecture (Fig.~\ref{fig:hand_design}(c)) with active flexion and passive elastic extension.

\paragraph*{Flexor tendon routing}
Finger flexion is driven by the HASEL stacks in the forearm. Each flexor tendon enters through the wrist mount, wraps around a pulley at the distal end of its actuator stack, and anchors to the tensioning gears at the base, doubling the tendon excursion relative to the actuator contraction (Sec.~\ref{subsec:actuator_design}) while maintaining a compact form factor.

\paragraph*{Extensor tendon routing}
Finger extension is achieved passively via an elastic return mechanism. 
The extensor tendons pass through a central hole in the wrist mount and connect in series with forearm elastic cords, which anchor to the bottom tensioning gears. 
The elastic restorative force is designed not to exceed the active flexion torque under maximum actuation.

To maintain structural stability during dynamic movements, each actuator stack is laterally reinforced: the sides of the actuators are sewn with fishing line, which is tensioned by serially connected elastic cables. As illustrated in Fig.~\ref{fig:actuator_overview}(d), this arrangement keeps the actuators aligned and prevents buckling during contraction.
All flexor and extensor tendons can be individually calibrated via the tensioning gears at the base of the forearm, allowing precise adjustment of resting tension and joint alignment.

\section{Control Framework}
\label{sec:control_framework}

\subsection{Self-Sensing Framework and Control Strategy}
\label{subsec:self_sensing_framework}

To eliminate the complexity, wiring, and weight that external sensors such as encoders and force transducers impose on multi-fingered hands, we implement a self-sensing feedback methodology that directly leverages the intrinsic electromechanical properties of the Peano-HASEL actuators.

\begin{figure}[tb]
\centering
\includegraphics[width=\columnwidth]{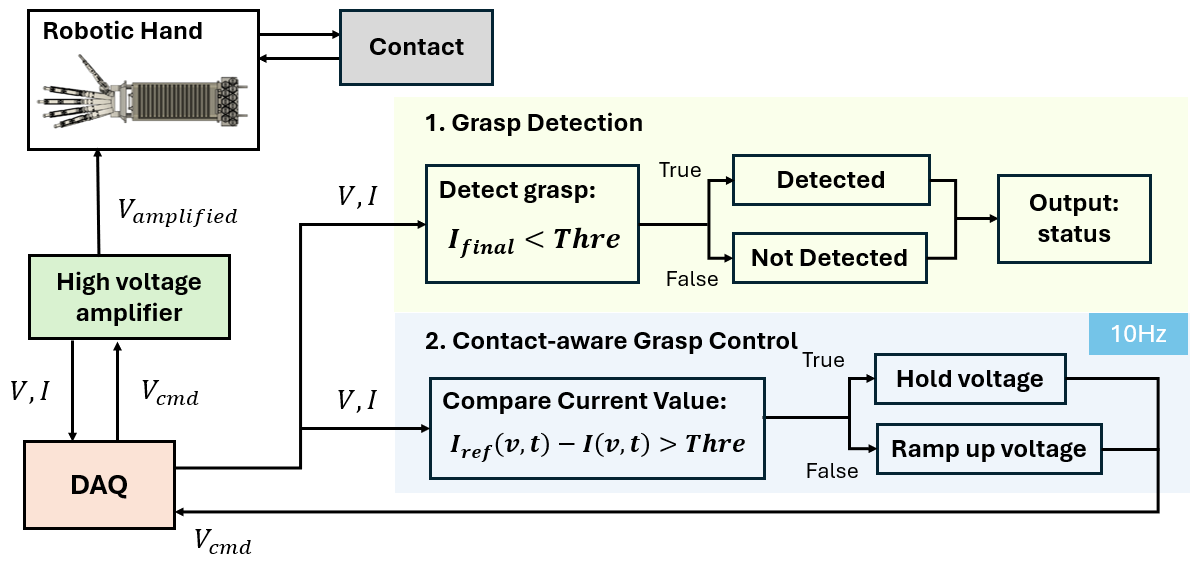}
\caption{\textbf{Overview of the current sensing feedback control mechanism.} The algorithms utilize real-time current measurement for Grasp Detection and Contact-Aware Grasp Control.}
\label{fig:self-sensing_algorithm}
\end{figure}

Composed of a liquid dielectric enclosed within flexible pouch electrodes, each actuator inherently functions as a variable capacitor. 
Unlike conventional methods that estimate state variations via secondary capacitance computations~\cite{hasel_shoulder, hasel_gripper, hasel_sensor}, we propose using the operating current as a direct, real-time indicator of kinematic activity.
Because the current profile is directly governed by the rate of geometric deformation, it can indicate contact without an intermediate capacitance estimate, reducing computational overhead and avoiding the noise amplification associated with numerical differentiation.
Consequently, a single low-pass-filtered current quantity enables both the grasp detection and contact-aware control algorithms summarized in Fig.~\ref{fig:self-sensing_algorithm}.

The operating current $i(t)$ of the HASEL actuator is governed by the first-order electromechanical model:
\begin{equation}
i(t) = C(t)\frac{dv(t)}{dt} + v(t)\frac{dC(t)}{dt}
\label{eq:current_model_base}
\end{equation}
In our framework, we apply a linear voltage ramp signal $v(t) = \alpha t$ (where $\alpha = 5.5\,\text{kV/s}$ for grasp detection and $\alpha = 6.0\,\text{kV/s}$ for contact-aware grasping), which yields a constant time derivative $\frac{dv(t)}{dt} = \alpha$. 
Substituting this condition into Equation~\eqref{eq:current_model_base} simplifies the total operating current to:
\begin{equation}
i(t) = \alpha C(t) + \alpha t\,\frac{dC(t)}{dt}
\label{eq:current_model}
\end{equation}
Here, the dynamic capacitance is geometrically modeled as 
$C(\beta) = \frac{\varepsilon_r \varepsilon_0 w}{2d} l_e(\beta)$ \cite{hasel_analytical_model},
where $\varepsilon_0$ is the vacuum permittivity, $\varepsilon_r$ is the relative permittivity of the shell material, $w$ is the width of the electrode, $d$ is the constant thickness of the individual shell layer, and $l_e(\beta)$ represents the effective active electrode length that changes with the central angle $\beta(t)$. 

During unconstrained free motion, the electrostatic Maxwell stress progressively zips the electrodes together, compressing the pouch and redistributing the dielectric fluid (increasing the effective active electrode length $l_e(\beta)$).
This redistribution leads to a smooth, monotonic capacitance increase ($\frac{dC(t)}{dt} > 0$), which constitutes a repeatable baseline current profile. Crucially, upon physical contact with an external object, the boundary condition restricts further geometric deformation, instantly halting the growth of the effective active electrode length $l_e(\beta)$.
Consequently, the rate of capacitance change sharply drops toward zero ($\frac{dC(t)}{dt} \rightarrow 0$), causing the second dynamic term in Equation~\eqref{eq:current_model} to vanish. 
This physical transition manifests as a distinct downward deviation in the current signal:
\begin{equation}
i_{\text{measured}}(t) \approx \alpha C_{\text{contact}} < i_{\text{free}}(t)
\end{equation}
where $C_{\text{contact}}$ is the capacitance at the moment of contact and $i_{\text{free}}$ is the free-motion baseline current.
This allows a current-based thresholding algorithm to directly map a mechanical boundary constraint to an electrical signal discontinuity without external sensors or capacitance estimation.

In this study, we monitor the operating current of the actuator stack driving the index MCP joint, which exhibited the most distinct and reliable current variations upon contact. 

\subsubsection{Grasp Detection Algorithm}
During each grasp attempt, the ramp voltage is applied, and the operating current is compared against a predefined threshold.
This threshold is selected based on pre-measured free-motion and successful-grasp profiles to ensure repeatable separation between the two states.  
Upon encountering a rigid object, the physical constraint halts deformation, causing the current to drop below the threshold. 
Notably, this current drop is triggered at the instant of initial contact, before any significant grasping force has developed.

\subsubsection{Contact-Aware Control for Delicate Grasping}
To manipulate fragile objects, a closed-loop control strategy continuously compares the operating current with a pre-recorded free-motion baseline trajectory. 
When the deviation between the measured and baseline currents exceeds a specified threshold, indicating physical contact, the controller immediately halts the voltage ramp and holds the voltage at its instantaneous value.
This strategy caps finger flexion upon contact, bounding the applied grasping force and preventing structural damage to the delicate object.

\section{Experimental Setup and Results}


\subsection{Experimental Setup and Robot Integration}
\label{subsec:experimental_setup_method}

To evaluate the hardware design and the theoretical control models, the fully assembled musculoskeletal hand was mounted on a Franka Research 3 arm (Franka Robotics GmbH) for dynamic grasping experiments. 
 
Data acquisition (DAQ) was executed at $1\,\text{kHz}$ via a National Instruments myDAQ system, while high-level control loops and threshold evaluations ran within custom MATLAB scripts at $10\,\text{Hz}$.
The low-voltage control signals generated by the DAQ system were amplified by a high-voltage amplifier (Trek 610C, Trek, Inc.) to provide driving voltages of up to $5.5\,\text{kV}$ to the Peano-HASEL actuator stacks, and up to $6.0\,\text{kV}$ exclusively for the contact-aware control tasks.

For current-sensing feedback, the applied voltage and operating current were measured via the Trek amplifier's built-in monitor channels. 
These analog signals were fed directly back to the DAQ system for threshold evaluation and baseline subtraction. 
A low-pass Butterworth filter was implemented to handle high-frequency electrical noise during static holding states.


\subsection{Characterization of Finger Performance} \label{subsec:characterization}
We evaluated the kinematic motion range and the static fingertip force of the index finger and thumb as a function of the applied voltage.

\subsubsection{Evaluation of Finger Motion Range}
To evaluate the operating range of the thumb and index finger, we measured joint angles relative to the applied voltage using AprilTags attached to each joint as position references. Finger motions were recorded at $30\,\text{fps}$ via a fixed camera during a $1\,\text{s}$ voltage ramp.

\begin{figure}[!tb]
\centering
\includegraphics[width=0.8\columnwidth]{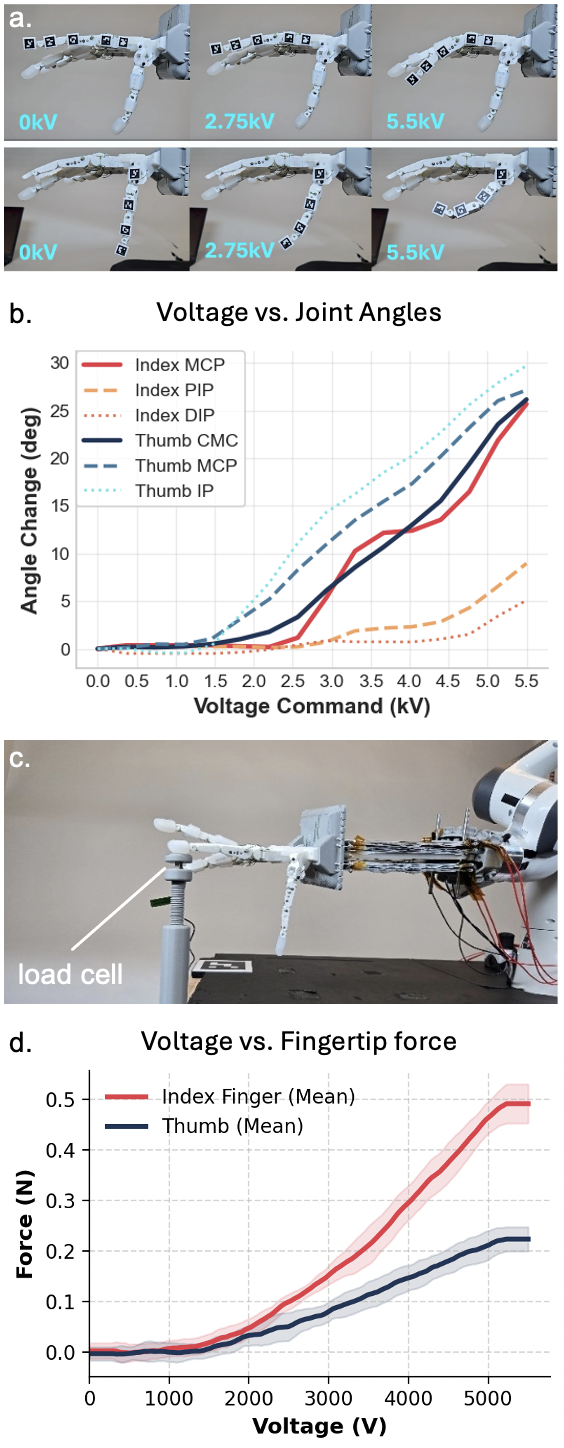}
\caption{\textbf{Characterization of the index finger and thumb performance.} (a, b) A $1\,\text{s}$ ramp signal was applied to the index finger and thumb, and the motion range of each joint was measured. (c, d) Under the same input conditions, the fingertip force was measured. Solid lines represent the mean value of five trials, with the shaded regions indicating the standard deviation.}
\label{fig:characterization}
\end{figure}

Figure~\ref{fig:characterization}(b) shows the voltage-angle relationship, with corresponding postures illustrated in Fig.~\ref{fig:characterization}(a).
Although both joint angles increase with voltage as expected, the fingers fail to achieve their theoretically expected full range of motion due to two practical limitations.

First, a deadband is observed in the low-voltage region. This is primarily attributed to the initial static friction in the tendon-routing system and to mechanical backlash at the joints, which together require a minimum driving force to initiate flexion.

Second, the maximum joint angle is limited to approximately $30^\circ$, falling well short of the intended $90^\circ$ flexion.
Beyond transmission losses from tendon-routing friction, we hypothesize that this constraint stems from a complex coupling effect caused by the high packing density of the actuator stacks within the human-sized forearm. 
Specifically, physical contact and volumetric bulging of adjacent contracting pouches increase inter-actuator mechanical resistance.
This dense configuration likely induces lateral forces and minor misalignments along the routing channels, further amplifying the routing friction itself. Consequently, the actuators reach their force limit under these interwoven friction sources before full flexion is achieved. Future iterations will require isolated friction analyses to experimentally validate these underlying factors.


\subsubsection{Evaluation of Fingertip Force}
We next measured the static fingertip force of the thumb and index finger using a load cell (DYLY-109, Daysensor) with a button on top, mounted on a vertical pole (Fig.~\ref{fig:characterization}(c)).
Each finger was initially set to a fully extended (straight) posture and positioned to press directly on the button.
For each digit, five independent trials were conducted by ramping the actuator voltage from $0$ to $5.5\,\text{kV}$ over $1\,\text{s}$ while recording the output force.

The resulting voltage-force relationship is plotted in Fig.~\ref{fig:characterization}(d).
At the maximum applied voltage of $5.5$\,kV, the index finger and thumb generated fingertip forces of $0.53$\,N and $0.26$\,N, respectively.
This variation is likely due to differences in tendon-routing paths, friction losses, and mechanical advantage.


To contextualize these metrics, the high-speed prosthetic finger in~\cite{hasel_finger} achieved a larger motion range ($77^\circ$) and forces exceeding $1\,\text{N}$ by driving a rigid four-bar linkage with parallel HASEL stacks (Table~\ref{tab:hasel_comparison}).
Our design instead prioritizes multi-fingered integration within a space-constrained, human-sized forearm, favoring a compliant tendon transmission and compact packaging over the maximized kinematics of an isolated joint.
These force levels align with our objective of inherent safety, proving sufficient for the stable, non-destructive grasping of lightweight objects.

\subsection{Compliance and Backdrivability} \label{subsec:compliance}
Compliance allows the hand to passively deform and mitigate excessive impact forces, ensuring safe interaction with humans and unstructured environments.
The proposed hand derives this inherent compliance from both the tendon transmission and the soft structure of the HASEL actuators. 

\begin{figure}[tb]
\centering
\includegraphics[width=0.8\columnwidth]{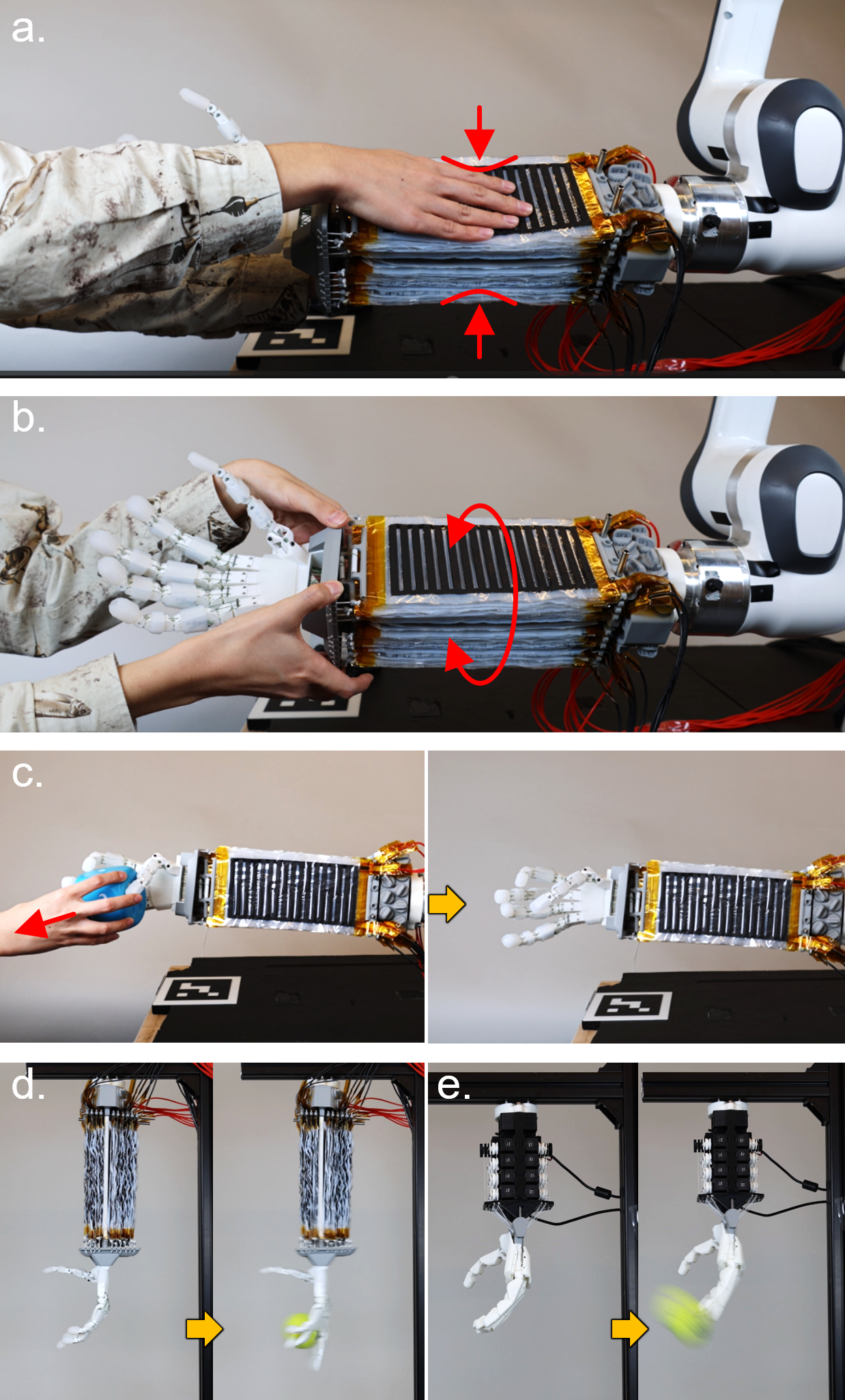}
\caption{\textbf{The proposed musculoskeletal hand system is inherently compliant.} (a, b) Demonstration of the mechanical compliance of the actuator packs in the forearm. The actuator packs deform when pushed by a human hand and accommodate twisting of the forearm. (c) The grasped object can be easily removed from the hand while constant voltage is applied, demonstrating its backdrivability. (d, e) Comparison of collision responses while actuators are active: the proposed HASEL-driven hand safely absorbs the impact of a tennis ball through large deformation, whereas a conventional motor-driven hand exhibits a rigid response.}
\label{fig:compliance}
\end{figure}

In the unpowered state, the entire system is highly flexible, with the forearm actuator packs deforming passively and accommodating twisting (Fig.~\ref{fig:compliance}(a, b)).

To evaluate active adaptability, we conducted an object-removal test under a constant holding voltage of $5.5\,\text{kV}$. 
When the grasped object was manually pulled, the fingers passively extended, allowing smooth removal without mechanism damage (Fig.~\ref{fig:compliance}(c)).

We further evaluated the dynamic backdrive behavior under sudden collisions.
A tennis ball was thrown at the fingers while they were actively flexed under a constant applied voltage.
As shown in Fig.~\ref{fig:compliance}(d), the proposed hand absorbed the impact by deforming backward and then returning to its initial flexed posture.
In contrast, a conventional motor-driven hand subjected to the same impact exhibited a rigid response with minimal deformation (Fig.~\ref{fig:compliance}(e)).


These results suggest that the system's inherent compliance and passive adaptability are well suited for physical human-robot collaboration.

\subsection{Versatile Grasping} \label{subsec:versatile_grasping}

To evaluate the practical grasping performance, we conducted grasping demonstrations across three representative grasp types from standard taxonomies~\cite{cutkosky1989grasp, feix2015grasp}: (1)~pinching (thumb and index finger), (2)~tripod grasp (thumb, index, and middle fingers), and (3)~power grasp (all fingers).
For each task, a $1\,\text{s}$ voltage ramp was applied to reach the target actuation level, followed by a constant holding voltage of $5.5$\,kV.

\begin{figure}[tb]
\centering
\includegraphics[width=\columnwidth]{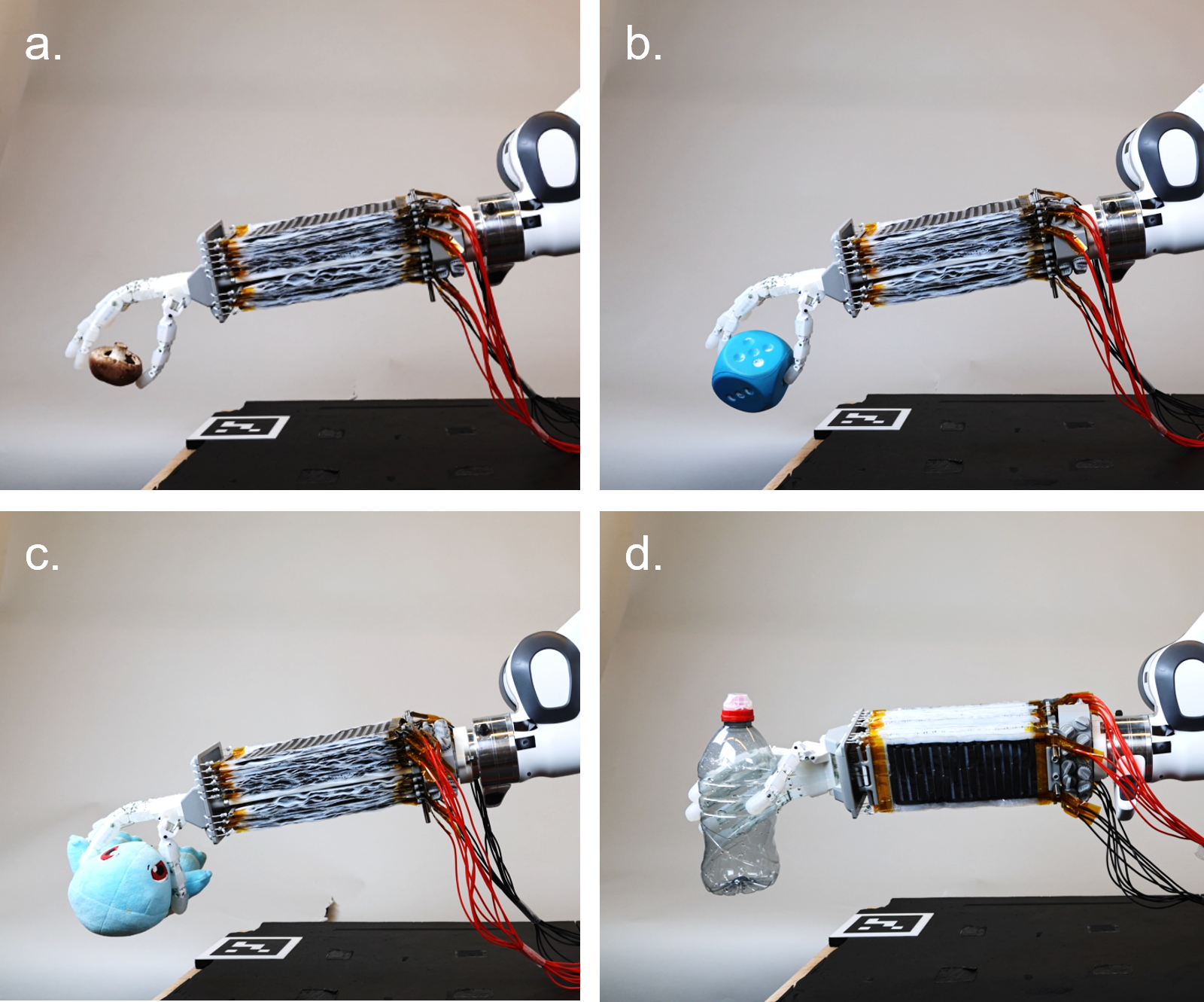}
\caption{\textbf{Demonstration of versatile grasp taxonomies adapting to different object properties.} (a, b) Pinching an 18 g mushroom and a 49 g cube using the thumb and index finger. (c) Tripod grasp of a 107 g soft stuffed toy. (d) Power grasp enveloping a 26 g empty PET bottle.}
\label{fig:grasp_taxonomy}
\end{figure}

As shown in Fig.~\ref{fig:grasp_taxonomy}, the hand successfully conformed to diverse object geometries. 
The hand achieved stable pinching of an $18$\,g mushroom and a $49$\,g cube (Fig.~\ref{fig:grasp_taxonomy}(a, b)), a tripod grasp on a $107$\,g soft stuffed toy (Fig.~\ref{fig:grasp_taxonomy}(c)), and a power grasp enveloping a $26$\,g empty PET bottle (Fig.~\ref{fig:grasp_taxonomy}(d)).
In all configurations, the fingers conformed to the object surfaces, maintaining stable grasps without noticeable slip or posture deviation.

These results show that the combination of multiple contact points and soft actuation enables stable grasping of lightweight objects across various geometries.

\subsection{Current-Sensing Feedback Control} \label{subsec:self_sensing}

\subsubsection{Grasp Detection}
\begin{figure}[tb]
  \centering
  \begin{subfigure}[b]{0.5\columnwidth}
    \centering
    \includegraphics[height=6.6cm, keepaspectratio]{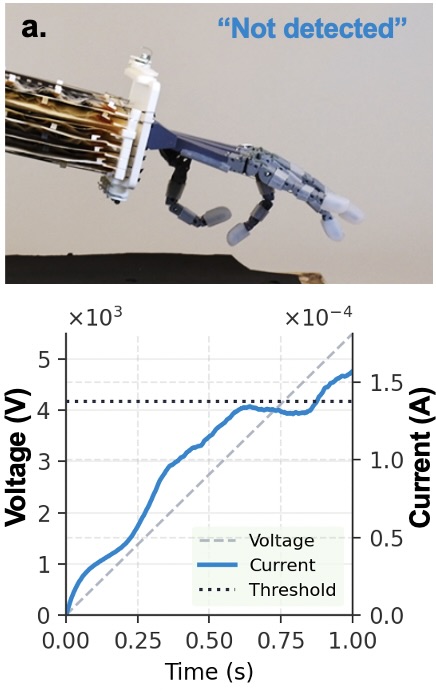}
    \caption{Not detected}
    \label{subfigure:not_detected}
  \end{subfigure}%
  \hfill
  \begin{subfigure}[b]{0.5\columnwidth}
    \centering
    \includegraphics[height=6.6cm, keepaspectratio]{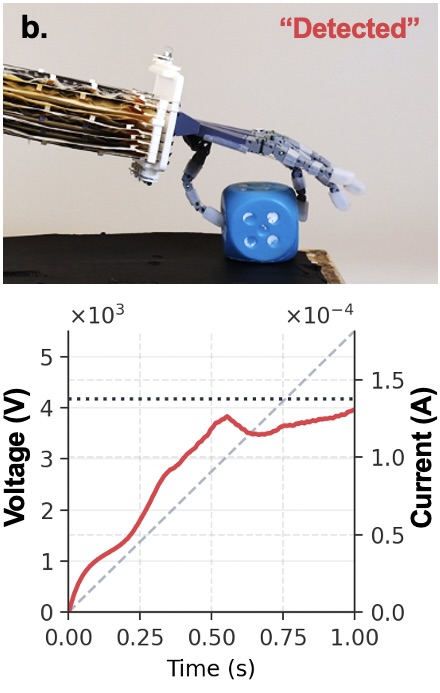}
    \caption{Detected}
    \label{subfigure:detected}
  \end{subfigure}
  
  \caption{\textbf{Results of the grasp detection experiment. }The current drops significantly when the finger motion is constrained by an object, allowing threshold-based detection.}
  \label{fig:grasp_detection}
\end{figure}

Experimental results (Fig.~\ref{fig:grasp_detection}) show that this threshold-based evaluation detects the presence of a grasped object purely from the electrical state of the actuator.


\subsubsection{Contact-Aware Control for Delicate Grasping}
\begin{figure}[tb]
  \centering
  \begin{subfigure}[b]{0.5\columnwidth}
    \centering
    \includegraphics[height=6.9cm, keepaspectratio]{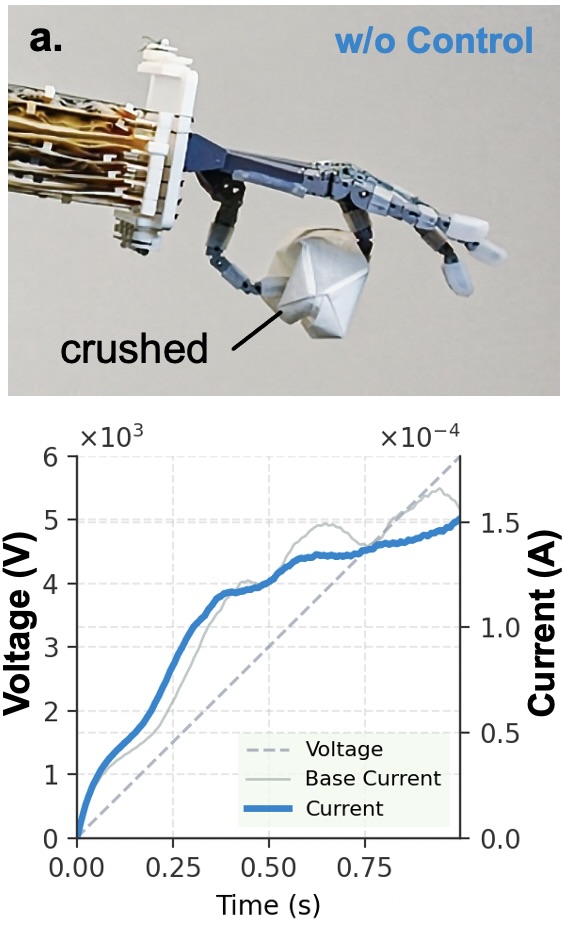}
    \caption{Without Control}
    \label{subfigure:wo_control}
  \end{subfigure}%
  \hfill
  \begin{subfigure}[b]{0.5\columnwidth}
    \centering
    \includegraphics[height=6.9cm, keepaspectratio]{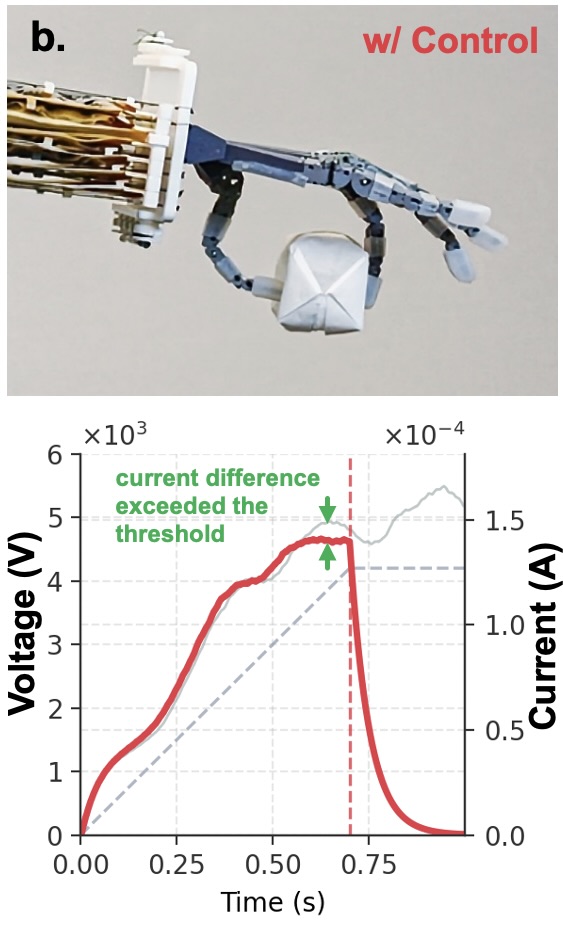}
    \caption{With Control}
    \label{subfigure:w_control}
  \end{subfigure}
  
  \caption{\textbf{Results of the contact-aware grasp control.} By holding the voltage when the current deviates from the free-motion baseline, the hand successfully grasps a fragile paper balloon without crushing it.}
  \label{fig:contact_aware_grasp}
\end{figure}

For the contact-aware grasping task, the maximum driving voltage was intentionally increased to $6.0\,\text{kV}$ (compared to $5.5\,\text{kV}$ in previous experiments) to amplify the dynamic current profile---both terms of Eq.~\eqref{eq:current_model} scale with the ramp rate $\alpha$---and clearly demonstrate the efficacy of the proposed current-based thresholding method.
As shown in Fig.~\ref{fig:contact_aware_grasp}, the controller halted the voltage ramp at the moment of contact and held the voltage, allowing the hand to grasp a fragile paper balloon without crushing it.

These results confirm that contact-aware manipulation and effective force limitation can be achieved by monitoring the actuator's operating current alone.


\section{Discussion}\label{sec:discussion}

To contextualize our contributions, Table~\ref{tab:hasel_comparison} provides a qualitative and quantitative comparison between the proposed hand and existing HASEL-driven systems, specifically the direct bending gripper~\cite{hasel_gripper} and the rigid-linkage prosthetic finger~\cite{hasel_finger}.

\begin{table*}[t]
\caption{Qualitative and Quantitative Comparison of HASEL-Driven Robotic Manipulators}
\label{tab:hasel_comparison}
\centering
\renewcommand{\arraystretch}{1.2}
\begin{tabular}{L{3.5cm} L{4cm} L{4cm} L{4.5cm}}
\toprule
\multicolumn{1}{l}{\textbf{Metrics}} & \textbf{Direct bending grippers \cite{hasel_gripper}} & \textbf{Single prosthetic finger \cite{hasel_finger}} & \textbf{Proposed musculoskeletal hand} \\
\midrule
\textbf{Kinematic form factor} & Simple pinching / wrapping gripper & Single isolated digit & Integrated anthropomorphic multi-fingered hand \\
\textbf{Actuator location} & Directly at the interface & Parallel stacks at the finger base & Remote forearm placement \\
\textbf{Transmission mechanism} & Direct deformation & Rigid four-bar linkage mechanism & 1:2 tendon-pulley amplification with rolling-contact joint \\
\textbf{Fingertip force} & $< 0.7$\,N & $> 1$\,N & $< 0.6$\,N \\
\textbf{Inherent compliance} & Very high (purely soft structure) & Moderate (rigid linkages) & High (tendons and soft muscles) \\
\textbf{Self-sensing \& feedback} & Capacitive self-sensing & N/A & Operating-current thresholding \\
\bottomrule
\end{tabular}
\end{table*}

Despite these promising results, the current system has several limitations, which future work will address as follows.

First, the fingers' maximum output force is limited and the effective range of motion is restricted to approximately $30^\circ$; as discussed in Sec.~\ref{subsec:characterization}, we hypothesize this arises from mechanical backlash and friction coupling among the densely packed actuator stacks and the tendon routing.
We will validate these bottlenecks through isolated friction analyses and address them with low-friction micro-channel guiding scaffolds, muscle architectures that minimize lateral bulging, and a synchronized antagonistic muscle system for preloading and torque--stiffness decoupling~\cite{kazemipour2025stretchable, kazemipour2026decoupling}.

Second, the passive elastic cords used for finger extension are susceptible to fatigue and creep over extended cycling; they will be replaced with metal springs for long-term durability.

Third, the threshold-based current-sensing algorithm lacks robustness against intense contact chattering and hysteresis-induced baseline shifts; adaptive thresholding, statistical filtering, and an empirical hysteresis compensation model will harden it.

Finally, the experimental evaluation remains preliminary: we will establish a statistical testing framework---multi-trial experiments across varied geometries, compliance levels, and heavier payloads---and benchmark repeatability, grasp stability, bandwidth, and energy efficiency against state-of-the-art motor-driven and soft robotic hands, paving the way toward dynamic in-hand manipulation.

\section{Conclusion}\label{sec:conclusion}
We presented an integrated musculoskeletal robotic hand powered entirely by remote Peano-HASEL actuators, combining a remote-actuation paradigm and a 1:2 pulley transmission with a compliant, human-inspired tendon-driven architecture to achieve a slim form factor and electrical isolation. Experiments demonstrated versatile grasping across standard taxonomies, while current-based self-sensing enabled grasp detection and closed-loop contact-aware control without external sensors, allowing the hand to delicately grasp highly fragile objects such as a paper balloon. These findings highlight the potential of HASEL-driven systems to combine physical safety and compliance with simplified mechanical and control architectures, positioning musculoskeletal electrohydraulic hands as a promising platform for safe, sensorless manipulation.





\section*{Acknowledgment}
This work was supported by the SNSF Project Grant \#200021\_215489. This support was instrumental in the development of this research.


\bibliographystyle{IEEEtran}
\bibliography{references}

\end{document}